\documentclass{article}
\PassOptionsToPackage{numbers,compress}{natbib}
\usepackage[preprint]{neurips_2026}
\usepackage[utf8]{inputenc}
\usepackage[T1]{fontenc}
\usepackage{newunicodechar}
\newunicodechar{ }{~}
\usepackage{hyperref,url,booktabs}
\hypersetup{
  hidelinks,
  pdftitle={DLWM: Diverse Latent World Models for Efficient Multimodal Reasoning},
  pdfauthor={David Huang and Lianlei Shan}
}
\usepackage{amsfonts,amssymb,amsmath}
\usepackage{nicefrac,microtype,xcolor}
\usepackage{algorithm,algorithmic}
\usepackage{enumitem}
\usepackage{graphicx,subcaption}
\title{DLWM: Diverse Latent World Models for Efficient Multimodal Reasoning}
\author{%
    David Huang \\ 
    University of Toronto \\ 
    \texttt{dawae.huang@mail.utoronto.ca} 
    \And Lianlei Shan \\ 
    Tsinghua University \\
    \texttt{shanlianlei18@ucas.edu.cn} 
}
\begin{document}
\maketitle
\vspace{-0.35in}
{\centering\small\textcolor{blue!45!black}{\href{https://dawae111.github.io/DLWM/}{\nolinkurl{dawae111.github.io/DLWM}}}\par}
\vspace{0.35in}
\begin{abstract}
Reasoning capabilities of multimodal large language models (MLLMs) have improved considerably in recent years. Existing approaches typically rely on explicit chain-of-thought or continuous latent-space trajectories to enhance multi-step reasoning. However, these methods generally assume that an input admits a single latent interpretation and unfold reasoning along a fixed path or under a uniform computation budget. In real-world multimodal settings, visual observations are often subject to occlusion, blur, viewpoint variation, or semantic ambiguity, giving rise to multiple plausible interpretations. A uniform reasoning strategy not only limits the model's ability to explore multiple hypotheses but also incurs high memory usage and rollout cost. We present DLWM (Diverse Latent World Models), a multimodal reasoning framework that combines latent-space reasoning with reinforcement learning. First, we construct a set of diverse latent world hypotheses in continuous latent space, each representing a candidate interpretation of the visual input, and unfold latent reasoning independently on each hypothesis. An orthogonality-based diversity regularizer discourages hypothesis collapse. Second, we formulate the latent reasoning process as a resource-constrained sequential decision problem and introduce a resource-aware reinforcement learning policy that adaptively allocates computation across hypotheses, dynamically deciding whether to expand a reasoning path, prune a hypothesis, merge reasoning paths, or produce the final answer, thereby substantially reducing memory footprint and improving rollout efficiency. Experiments on multiple multimodal reasoning benchmarks demonstrate that DLWM outperforms existing methods by 2--5 points in accuracy while reducing memory usage by 24\%.

\end{abstract}

\section{Introduction}
\label{sec:intro}

Multimodal large language models have advanced rapidly in visual question answering, visual reasoning, and complex decision-making tasks. Compared with pure text reasoning, however, multimodal reasoning faces a more fundamental challenge: \emph{perceptual uncertainty}. In the real world, visual inputs are frequently affected by occlusion, blur, viewpoint changes, and semantic ambiguity, so that a single observation may correspond to multiple plausible interpretations. Most existing methods compress multimodal inputs into a single semantic representation and reason over it---implicitly assuming a unique ground-truth world. This assumption often breaks down in complex visual scenes, limiting robustness and generalization under uncertainty. At the same time, as the number of reasoning paths grows, computation and memory costs escalate rapidly, making the trade-off between performance and efficiency an equally pressing concern.

\begin{figure}[t]
  \centering
  \includegraphics[width=\linewidth]{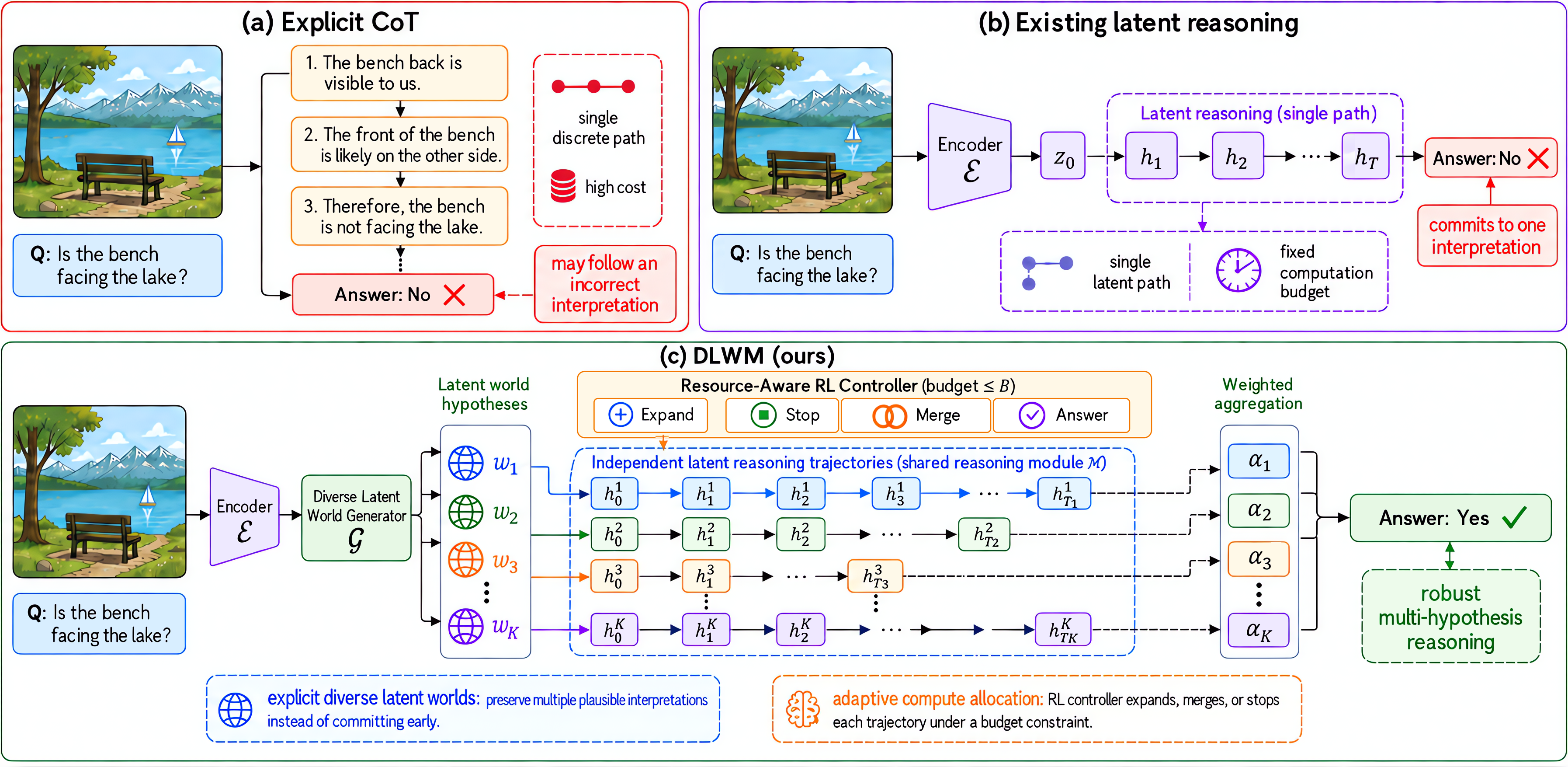}
  \caption{
  Comparison of prior multimodal reasoning methods and our proposed DLWM framework.
  Explicit chain-of-thought methods reason through a single textual path, while existing latent reasoning methods typically rely on a single latent state or fixed rollout process.
  In contrast, DLWM constructs multiple diverse latent world hypotheses and uses a resource-aware RL controller to adaptively expand, stop, merge, or answer across reasoning paths.
  }
  \label{fig:intro_comparison}
\end{figure}

Existing multimodal reasoning methods fall into two broad categories: explicit chain-of-thought (CoT) methods and continuous latent-space reasoning methods. CoT methods improve multi-step reasoning by generating intermediate textual steps~\citep{wei2022chain, kojima2022large}, but the reasoning process operates in discrete token space and consumes substantial computation to maintain linguistic coherence~\citep{madaan2022text}. Continuous latent-space methods such as Coconut~\citep{hao2024training} and CoLaR~\citep{tan2025think} shift reasoning into hidden states, improving efficiency, yet their multi-path capability exists only in superposition and lacks an explicit, controllable multi-hypothesis structure. In the multimodal domain, LVR~\citep{li2025latent}, LaRe~\citep{ma2025multimodal}, and Heima~\citep{shen2025efficient} extend latent reasoning to vision-language tasks but still rely on single-path reasoning or fixed-structure iterative processes with no mechanism for dynamic selection among paths. Methods that explore dynamic control of reasoning, such as Soft Thinking~\citep{zhang2025soft} and Overclocking~\citep{eisenstadt2025overclocking}, modulate latent representations continuously but lack discrete structural operations---expansion, termination, or merging---across different paths.

To address these problems, we present \textbf{DLWM} (Diverse Latent World Models), a multimodal reasoning framework that unifies multi-hypothesis latent modeling with resource-aware reinforcement learning. The core idea is to explicitly construct multiple \emph{latent world hypotheses} in continuous latent space, each intended to represent one possible interpretation of the input, thereby enabling the model to unfold reasoning along multiple semantic paths in parallel. To reduce representational collapse among hypotheses, we introduce an orthogonality-based diversity regularizer that encourages representational differences. Building on this, we further formulate the latent reasoning process as a budget-constrained sequential decision problem and introduce a resource-aware RL policy that dynamically allocates computation across latent worlds. The policy adaptively decides whether to expand, terminate, or merge reasoning paths, or to directly output an answer, thereby preserving accuracy while reducing redundant computation.

Our main contributions are as follows:
\begin{itemize}[nosep,leftmargin=1.5em]
  \item We introduce an explicit \emph{diverse latent world hypothesis} mechanism that constructs multiple candidate interpretations of the visual input in continuous space, with orthogonal regularization to reduce representational collapse.
  \item We propose a \emph{resource-aware reinforcement learning} compute allocation mechanism that formalizes latent reasoning as a resource-constrained sequential decision problem, enabling dynamic scheduling of reasoning computation while reducing overhead.
  \item We unify multi-hypothesis modeling and efficient reasoning within a single framework, alleviating the tension between expressiveness and computational cost in multi-path reasoning.
  \item Experiments on multiple multimodal reasoning benchmarks demonstrate improved accuracy and an improved accuracy--efficiency trade-off over the compared approaches.
\end{itemize}

\newpage
\section{Related Work}
\label{sec:related}

\subsection{Latent-Space Reasoning}
\label{sec:related_latent}

Chain-of-thought (CoT) methods have significantly improved multi-step reasoning in large language models~\citep{wei2022chain, kojima2022large}, but their reliance on discrete token generation consumes substantial computation to maintain linguistic coherence, limiting reasoning efficiency~\citep{madaan2022text}. To mitigate this bottleneck, recent work shifts reasoning from explicit text space into continuous latent space, completing intermediate ``thinking'' in hidden states to reduce redundant text generation and improve efficiency.
In the text domain, Coconut~\citep{hao2024training} and CoLaR~\citep{tan2025think} have shown that latent-space reasoning can compress intermediate processes while preserving reasoning capability; Soft Thinking~\citep{zhang2025soft}, SEAL~\citep{chen2025seal}, and Overclocking~\citep{eisenstadt2025overclocking} further explore continuous control over latent reasoning and reasoning-depth modulation. These methods demonstrate that continuous latent space offers more flexible reasoning representations and control than discrete CoT. However, most existing methods still operate along a single reasoning path---even when some implicitly encode multiple possible trajectories in the latent space, they lack explicit modeling and structured comparison of different hypotheses.

In the multimodal setting, LVR~\citep{li2025latent}, LaRe~\citep{ma2025multimodal}, and Heima~\citep{shen2025efficient} extend latent reasoning to vision-language tasks, making progress in perceptual modeling and reasoning efficiency. Yet these methods still primarily rely on single latent representations or single-path rollouts, lacking explicit modeling of multiple plausible interpretations under the same visual input, and thus tend to prematurely converge to a single (possibly incorrect) interpretation in the presence of visual ambiguity or perceptual uncertainty.
Unlike prior work, DLWM explicitly constructs multiple latent world hypotheses in continuous space and encourages representational diversity among hypotheses via orthogonal regularization, enhancing the model's ability to express and explore multiple candidate interpretations. InsTALL~\citep{nguyen2025installcontextawareinstructionaltask} uses task graphs for instructional assistance over visual streams; it is related through structured intermediate representations, but does not study latent multi-hypothesis allocation.

\subsection{Reinforcement Learning and Adaptive Computation}
\label{sec:related_rl}

Early applications of RL to language models focused on alignment via human feedback~\citep{ouyang2022training}, optimizing output-level preference rather than modeling intermediate reasoning decisions, making them unsuitable for fine-grained control of reasoning paths or computation allocation.
Recent work treats the reasoning process itself as a sequential decision problem. CoLaR~\citep{tan2025think} uses RL to optimize latent chain length, learning shorter yet correct paths, but addresses only single-path compression without multi-path allocation. Methods for reasoning speed control and representation manipulation~\citep{chen2025seal, eisenstadt2025overclocking} attempt to modulate reasoning depth through continuous signals in latent space. These approaches show that RL and reasoning control can improve efficiency, but most target single-path reasoning, addressing ``how long to reason'' rather than ``how to allocate computation across multiple paths.''
A related line of work on adaptive computation time~\citep{graves2016adaptive} and dynamic computation allocation shows that different samples should receive different reasoning budgets for a better performance--efficiency trade-off. However, these methods still operate along single trajectories without explicit multi-path modeling, and cannot support path-level operations such as expansion, termination, or merging of reasoning branches.
In contrast, DLWM formulates multimodal latent reasoning as a resource-constrained sequential decision problem, where a resource-aware RL policy dynamically allocates computation across multiple latent worlds. The policy controls not only reasoning depth along individual paths but also executes structural decisions (expand, stop, merge) across hypotheses, unifying multi-hypothesis exploration with computation efficiency optimization.

\section{Method}
\label{sec:method}

\begin{figure}[t]
  \centering
  \includegraphics[width=\linewidth]{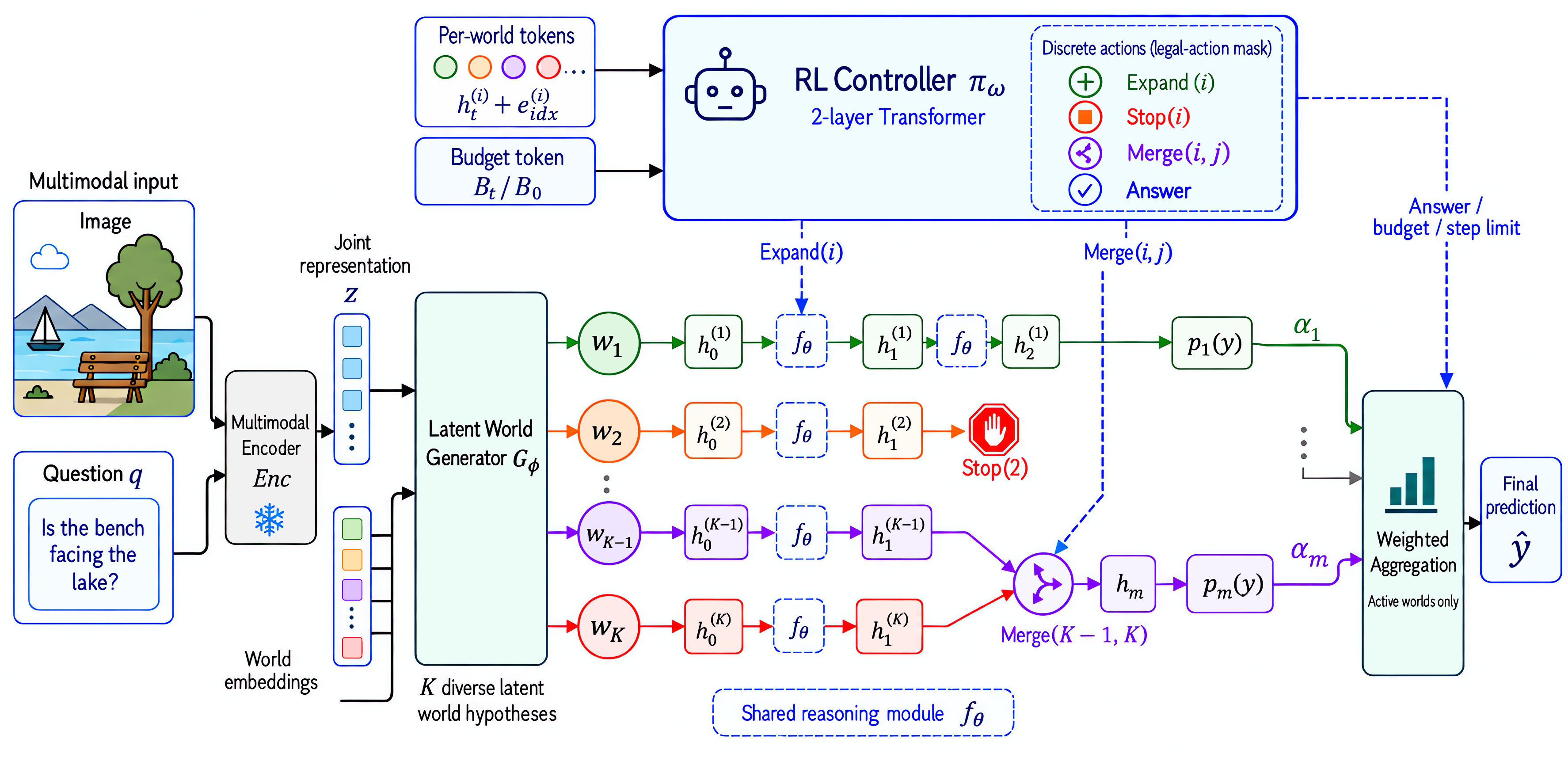}
  \caption{Overview of \textbf{DLWM}. Given a multimodal input, the encoder produces a joint representation $\mathbf{z}$, from which the Latent World Generator creates $K$ diverse world hypotheses. Each world initializes an independent reasoning trajectory updated by a shared reasoning module. An RL controller dynamically allocates computation by selecting \textsc{Expand}, \textsc{Stop}, \textsc{Merge}, or \textsc{Answer} actions under a budget constraint. Active worlds are aggregated via learned weighted fusion to produce the final prediction.}
  \label{fig:overview}
\end{figure}

We propose \textbf{DLWM} (Diverse Latent World Models), which explicitly constructs multiple latent world hypotheses in continuous space and dynamically allocates computation across them via a resource-aware RL policy. DLWM consists of (1) a \emph{Diverse Latent World Hypothesis Generator} that produces $K$ candidate representations of the input, and (2) a \emph{Resource-Aware RL Controller} that adaptively schedules computation across their reasoning trajectories under a budget constraint. An overview is shown in Figure~\ref{fig:overview}.

\subsection{Diverse Latent World Hypotheses}
\label{sec:worlds}

\paragraph{Multimodal Encoding.}
Given image $I$ and question $q$, a vision-language encoder produces a joint representation:
\begin{equation}
  \mathbf{z} = \mathrm{Enc}(I, q), \quad \mathbf{z} \in \mathbb{R}^d.
  \label{eq:encoder}
\end{equation}
The encoder can be any pretrained multimodal backbone (\emph{e.g.}, the vision-language tower of a MLLM). We keep it frozen throughout training.

\paragraph{Latent World Generation.}
We introduce a latent world generator $G_\phi$ that produces $K$ world representations from the shared encoding. Each world is conditioned on a learnable \emph{world embedding} $\mathbf{e}_i \in \mathbb{R}^d$, which serves as a hypothesis identifier guiding the model to interpret the same input from different perspectives:
\begin{equation}
  \mathbf{w}_i = G_\phi(\mathbf{z}, \mathbf{e}_i), \quad i = 1, \ldots, K.
  \label{eq:world_gen}
\end{equation}
The world embeddings are initialized near-orthogonally using Gram--Schmidt and updated by the task and orthogonality losses, without separate labels.
The resulting world representations are intended to capture different candidate interpretations, all grounded in the same input $\mathbf{z}$. Different worlds share the same latent reasoning module (Section~\ref{sec:rl}), ensuring that reasoning trajectories are comparable while avoiding parameter proliferation.

\paragraph{Latent Reasoning.}
Each world representation initializes an independent reasoning trajectory:
\begin{equation}
  \mathbf{h}_0^{(i)} = \mathbf{w}_i, \qquad
  \mathbf{h}_{t+1}^{(i)} = f_\theta\!\left(\mathbf{h}_t^{(i)}\right),
  \label{eq:reasoning}
\end{equation}
where $f_\theta$ is a shared latent reasoning module implemented as six Transformer blocks with tied weights across worlds and time steps. This weight-sharing provides an inductive bias toward iterative refinement~\citep{saunshi2025reasoning} while keeping the reasoning-module parameter count independent of $K$. This update is applied only when $\texttt{Expand}(i)$ is selected; other retained world states remain unchanged.

\paragraph{World Diversity Regularization.}
Without explicit encouragement, the generator may collapse all world representations to the same point, making the $K$ trajectories redundant. We mitigate this with an orthogonality loss that penalizes directional similarity between world representations:
\begin{equation}
  \mathcal{L}_{\mathrm{orth}} = \sum_{i \neq j} \left(\frac{\mathbf{w}_i^\top \mathbf{w}_j}{\lVert \mathbf{w}_i \rVert \, \lVert \mathbf{w}_j \rVert}\right)^2.
  \label{eq:orth}
\end{equation}
\paragraph{Multi-World Aggregation.}
After reasoning terminates (see Section~\ref{sec:rl}), each active world $i \in \mathcal{A}$ produces a task prediction:
\begin{equation}
  p_i(y) = p(y \mid \mathbf{h}_T^{(i)}).
\end{equation}
The final output is a weighted mixture:
\begin{equation}
  p(y \mid I, q) = \sum_{i \in \mathcal{A}} \alpha_i \, p_i(y), \quad
  \alpha_i = \mathrm{softmax}\!\left(s_\psi(\mathbf{h}_T^{(i)})\right),
  \label{eq:aggregate}
\end{equation}
where $s_\psi$ is a lightweight scoring network that evaluates the final reasoning state of each world; the softmax is normalized over the active set $\mathcal{A}$.
Orthogonality does not guarantee semantically distinct interpretations; we examine answer disagreement, collapse rates, and oracle coverage in Appendix~\ref{app:diversity_diagnostics}.

\subsection{Resource-Aware RL for Compute Allocation}
\label{sec:rl}

Multi-world modeling enhances the model's ability to represent perceptual uncertainty, but it also introduces additional reasoning overhead. If all latent worlds execute fixed-length rollouts, total computation grows linearly with the number of worlds and reasoning depth (Appendix~\ref{app:complexity}). We frame the reasoning process as a budget-constrained Markov decision process and train an RL controller to allocate compute dynamically.

\begin{algorithm}[!tb]
\caption{Resource-Aware Dynamic Reasoning Training of DLWM}
\label{alg:dlwm}
\begin{algorithmic}[1]
\REQUIRE Training set $\mathcal{D}$, encoder $\mathrm{Enc}$ (frozen), world generator $G_\phi$, reasoning module $f_\theta$, scoring head $s_\psi$, RL controller $\pi_\omega$, worlds $K$, max controller steps $T_{\max}$, budget $B$
\ENSURE Trained parameters $\{\phi, \theta, \psi, \omega, \{\mathbf{e}_i\}_{i=1}^K\}$
\STATE Initialize parameters $\phi, \theta, \psi, \omega$ and world embeddings $\{\mathbf{e}_i\}_{i=1}^K$
\STATE Supervised pretraining of $G_\phi,f_\theta,s_\psi$ and world embeddings at fixed rollout length
\FOR{each RL fine-tuning iteration}
  \STATE Sample minibatch $\{(I, q, y)\} \sim \mathcal{D}$
  \FOR{each $(I, q, y)$}
    \STATE $\mathbf{z} \gets \mathrm{Enc}(I, q)$
    \STATE $\{\mathbf{w}_i\}_{i=1}^K \gets \{G_\phi(\mathbf{z}, \mathbf{e}_i)\}_{i=1}^K$ \hfill \COMMENT{Generate diverse worlds}
    \STATE $\mathbf{h}_0^{(i)} \gets \mathbf{w}_i$, \; $\mathcal{A} \gets \{1, \ldots, K\}$, \; $t \gets 0$, \; $B_t \gets B$
    \WHILE{$t < T_{\max}$ \textbf{and} $B_t > 0$ \textbf{and} $\mathcal{A} \neq \varnothing$}
      \STATE Build per-world/index tokens and a budget token from the current active states
      \STATE Mask invalid actions, including Stop on the last active world
      \STATE $a_t \sim \pi_\omega(\cdot \mid s_t)$ \hfill \COMMENT{RL controller selects action}
      \STATE $\mathbf{h}_{t+1}^{(k)} \gets \mathbf{h}_t^{(k)}$ for all $k \in \mathcal{A}$ \hfill \COMMENT{Retain unmodified states}
      \IF{$a_t = \texttt{Expand}(i)$}
        \STATE $\mathbf{h}_{t+1}^{(i)} \gets f_\theta(\mathbf{h}_t^{(i)})$; \; $B_t \gets B_t - c_{\mathrm{exp}}$
      \ELSIF{$a_t = \texttt{Stop}(i)$}
        \STATE $\mathcal{A} \gets \mathcal{A} \setminus \{i\}$; \; $B_t \gets B_t - c_{\mathrm{stop}}$
      \ELSIF{$a_t = \texttt{Merge}(i,j)$}
        \STATE $\tilde{\mathbf{h}}_{t+1} \gets \tfrac12\mathbf{h}_t^{(i)} + \tfrac12\mathbf{h}_t^{(j)}$; replace $i,j$ by this merged world in $\mathcal{A}$; $B_t \gets B_t - c_{\mathrm{merge}}$
      \ELSIF{$a_t = \texttt{Answer}$}
        \STATE $t \gets t + 1$; \textbf{break}
      \ENDIF
      \STATE $t \gets t + 1$
    \ENDWHILE
    \STATE $T \gets t$; compute $p(y \mid I, q)$ via Eq.~\eqref{eq:aggregate} \hfill \COMMENT{Aggregate active worlds}
\STATE $\mathcal{L}_{\mathrm{task}} \gets -\log p(y \mid I, q)$ \hfill \COMMENT{Supervised objective}
\STATE $\mathcal{L}_{\mathrm{orth}} \gets \sum_{i \neq j} \bigl(\mathbf{w}_i^\top \mathbf{w}_j / (\lVert\mathbf{w}_i\rVert\lVert\mathbf{w}_j\rVert)\bigr)^2$
\STATE $R \gets R_{\mathrm{acc}} - \lambda_T T - \lambda_K \bar{K}$ \hfill \COMMENT{Reward signal}
  \ENDFOR
  \STATE Update $\phi, \theta, \psi, \{\mathbf{e}_i\}_{i=1}^K$ on $\mathcal{L}_{\mathrm{task}} + \lambda_{\mathrm{orth}} \mathcal{L}_{\mathrm{orth}}$ \hfill \COMMENT{Supervised objective update}
  \STATE Update $\omega$ via policy gradient on $\mathcal{L}_{\mathrm{RL}}$
\ENDFOR
\STATE \textbf{return} $\{\phi, \theta, \psi, \omega, \{\mathbf{e}_i\}_{i=1}^K\}$
\end{algorithmic}
\end{algorithm}

\paragraph{MDP Formulation.}
At each step $t$, the \textbf{state} consists of the active world representations and the remaining budget:
\begin{equation}
  s_t = \left(\left\{\mathbf{h}_t^{(i)}\right\}_{i \in \mathcal{A}_t},\; B_t\right),
  \label{eq:state}
\end{equation}
where $\mathcal{A}_t$ is the set of active worlds and $B_t \in \mathbb{R}_{\geq 0}$ is the remaining compute budget. For the controller's input, each active world contributes a token containing its state and a learned world-index embedding; a separate token encodes $B_t/B_0$. A two-layer Transformer contextualizes these tokens. Expand and Stop produce one logit per world; Merge produces one per unordered pair using token concatenation, difference, and element-wise product; Answer reads the contextualized budget token. A single softmax normalizes the legal action logits after masking.

The controller $\pi_\omega(a \mid s_t)$ selects an \textbf{action} from:
\begin{equation}
  a_t \in \bigl\{\text{\texttt{Expand}}(i),\; \text{\texttt{Stop}}(i),\; \text{\texttt{Merge}}(i,j),\; \text{\texttt{Answer}}\bigr\},
  \label{eq:actions}
\end{equation}
with the following effects and costs:
\begin{itemize}[nosep,leftmargin=1.5em]
  \item $\texttt{Expand}(i)$: advance world $i$ by one reasoning step, $\mathbf{h}_{t+1}^{(i)} = f_\theta(\mathbf{h}_t^{(i)})$. Cost: $c_{\mathrm{exp}}$.
  \item $\texttt{Stop}(i)$: remove world $i$ from the active set $\mathcal{A}$ and final aggregation. This action is masked when only one world remains. Cost: $c_{\mathrm{stop}}$.
  \item $\texttt{Merge}(i,j)$: combine two selected worlds into one:
  \begin{equation}
    \tilde{\mathbf{h}}_t = \beta \, \mathbf{h}_t^{(i)} + (1-\beta) \, \mathbf{h}_t^{(j)},
    \label{eq:merge}
  \end{equation}
  where $\beta=0.5$ by default, thereby reducing the number of parallel reasoning paths. The merged world replaces both parents in $\mathcal{A}$. Cost: $c_{\mathrm{merge}}$.
  \item $\texttt{Answer}$: terminate reasoning and produce the final prediction via Eq.~\eqref{eq:aggregate}. Cost: $0$.
\end{itemize}
Reasoning also terminates if $B_t \leq 0$ or $t$ reaches the controller-step limit $T_{\max}$, using the current active worlds for aggregation. Worlds not selected for further expansion remain active unless stopped or merged. Stop consumes budget for another controller transition, whereas Answer ends the episode; late answers are penalized by $\lambda_T T$.

\paragraph{Reward Design.}
The reward jointly encourages task accuracy and computational efficiency:
\begin{equation}
  R = R_{\mathrm{acc}} - \lambda_T \, T - \lambda_K \, \bar{K},
  \label{eq:reward}
\end{equation}
where $R_{\mathrm{acc}}$ is a rule-based correctness reward (\emph{e.g.}, exact match), $T$ is the total number of controller steps taken, and $\bar{K}$ is the average number of active worlds across steps. This design incentivizes the controller to terminate early on easy instances while preserving multiple worlds for ambiguous ones. The defaults are $\lambda_T=0.05$, $\lambda_K=0.10$, $c_{\mathrm{exp}}=1.0$, $c_{\mathrm{stop}}=0.1$, and $c_{\mathrm{merge}}=0.5$. They were selected by one-factor-at-a-time sweeps on held-out validation data, without using MMVP for selection. Sensitivity and stopped-world analyses are in Appendix~\ref{app:controller_diagnostics}.

\paragraph{Policy Optimization.}
The policy network $\pi_\omega(a \mid s)$ is optimized via policy gradient. The optimization objective is:
\begin{equation}
  J(\omega) = \mathbb{E}_{\pi_\omega}[R],
  \label{eq:policy_obj}
\end{equation}
For a trajectory $\tau$ with $T$ actions and a moving-average baseline $b$:
\begin{equation}
  \nabla_\omega J(\omega) = \mathbb{E}_{\tau\sim\pi_\omega}\!\left[\sum_{t=0}^{T-1}\nabla_\omega \log \pi_\omega(a_t \mid s_t) \cdot (R-b)\right].
  \label{eq:policy_grad}
\end{equation}

\paragraph{Training Pipeline.}
We train the generator $G_\phi$, reasoning module $f_\theta$, and scoring head $s_\psi$ jointly on $\mathcal{L}_{\mathrm{task}} + \lambda_{\mathrm{orth}} \mathcal{L}_{\mathrm{orth}}$, where $\mathcal{L}_{\mathrm{task}} = -\log p(y \mid I, q)$, and update the controller $\pi_\omega$ separately via policy gradient on $\mathcal{L}_{\mathrm{RL}} = -\mathbb{E}_{\pi_\omega}[R]$. This decoupling prevents RL gradient noise from destabilizing representation learning. Since RL training from scratch is often unstable, we adopt a two-stage strategy: (i)~supervised pretraining at fixed rollout length to learn stable representations and base reasoning, followed by (ii)~RL fine-tuning to optimize compute allocation. The full procedure is given in Algorithm~\ref{alg:dlwm}; Appendix~\ref{app:algorithm} provides a worked example.

\section{Experiments}
\label{sec:experiments}

\subsection{Experimental Setup}
\label{sec:setup}

\paragraph{Datasets.}
We evaluate DLWM on four multimodal reasoning benchmarks covering perception-intensive, compositional, open-domain, and cross-modal reasoning tasks. 
\textbf{MMVP}~\citep{tong2024eyeswideshutexploring} emphasizes fine-grained visual perception under visual confusability. 
\textbf{GQA}~\citep{hudson2018gqa} focuses on compositional and multi-hop relational reasoning over real images. 
\textbf{VQAv2}~\citep{balanced_vqa_v2} contains diverse everyday scenes with strong language priors, testing robust vision-language fusion. 
\textbf{ScienceQA}~\citep{lu2022learn} combines visual, textual, and scientific knowledge, requiring multi-step cross-modal reasoning.
We train one model on the union of the GQA, VQAv2, and ScienceQA training sets; MMVP is strictly zero-shot and is not used for training, validation, or hyperparameter selection. To address language-prior and visual-dependency limitations, we also evaluate six newer benchmarks zero-shot (Table~\ref{tab:new_benchmarks}).

\paragraph{Implementation Details.}
DLWM is built on Qwen2.5-VL-7B-Instruct~\citep{bai2025qwen25vltechnicalreport} with its native vision tower and language backbone frozen. 
The Latent World Generator is a two-layer MLP with approximately orthogonal learnable world embeddings, using $K{=}4$ worlds by default. 
The latent reasoning module contains 6 shared Transformer blocks with hidden size 4096, Pre-LN, and GELU activations. 
A single-layer MLP is used for world scoring, and a two-layer Transformer encoder with hidden size 1024 and 8 attention heads serves as the RL policy network. The added trainable modules contain 126M parameters (approximately 1.8\% of the 7B backbone), including 36M for the controller.
We set the initial budget to $B{=}10$ and maximum number of controller steps to $T_{\max}{=}8$. 
Training follows a two-stage strategy: supervised pretraining with fixed rollout length, followed by RL fine-tuning using REINFORCE with a moving-average baseline. 
We use AdamW with learning rate $2 \times 10^{-5}$, batch size 64, mixed precision, and 8 A100 GPUs. 
By default, at inference we use greedy action selection with at most 4 parallel worlds.
Appendix~\ref{app:implementation} gives the evaluation and measurement protocols.

\paragraph{Baselines and Evaluation Metrics.}
We compare with explicit \textbf{CoT}~\citep{wei2022chain}, latent reasoning methods \textbf{Coconut}~\citep{hao2024training} and \textbf{CoLaR}~\citep{tan2025think}, multimodal latent reasoning methods \textbf{LVR}~\citep{li2025latent}, \textbf{LaRe}~\citep{ma2025multimodal}, and \textbf{Heima}~\citep{shen2025efficient}, as well as the inference-time latent method \textbf{Soft Thinking}~\citep{zhang2025soft}. 
All baselines use the same frozen backbone, input resolution, and training data, and are evaluated under the same maximum compute budget. We use released method-specific hyperparameters where available and shared values otherwise.

We report task accuracy for reasoning performance, using standard accuracy for GQA, ScienceQA, and MMVP, and consensus-based accuracy for VQAv2. 
For efficiency, we report average rollout steps, average number of active worlds, GPU memory usage, and inference throughput. Controller decisions and world updates are counted separately. Table~\ref{tab:main} reports three-seed means for DLWM, LVR, and LaRe; other methods retain single-run results. Full statistics are in Appendix~\ref{app:seed_results}.

\subsection{Main Results}
\label{sec:main_results}

DLWM achieves the highest accuracy on all four benchmarks (Table~\ref{tab:main}). Over the strongest multimodal latent reasoning baselines (LVR and LaRe), DLWM gains 2.4--2.9 points on GQA and 2.6--3.1 on ScienceQA---tasks that require multi-step reasoning and cross-modal integration---indicating that explicit multi-hypothesis modeling enables more thorough exploration across semantic interpretations and reduces error accumulation from single-path commitment.

The largest gains appear on MMVP (+5.2 over LVR, +4.4 over LaRe), which targets visual confusability. This supports the value of multi-hypothesis modeling: by maintaining multiple latent interpretations simultaneously, the model can avoid premature convergence to an incorrect hypothesis and dynamically filter and fuse worlds during reasoning. In contrast, Soft Thinking implicitly encodes multi-path information in latent space but lacks explicit path structure.

Compared with Coconut, DLWM achieves higher accuracy on all four benchmarks. Relative to CoLaR, which uses RL to compress a single path, DLWM allocates computation \emph{across} multiple paths, achieving a better performance--efficiency balance.

\begin{table*}[t]
\caption{Main benchmark results and efficiency comparison. Best results are in \textbf{bold}.}
\label{tab:main_efficiency}
\centering
\small
\renewcommand{\arraystretch}{0.95}

\begin{subtable}[t]{0.48\textwidth}
\centering
\caption{Main results (Acc. \%).}
\label{tab:main}
{\setlength{\tabcolsep}{3.2pt}
\resizebox{\linewidth}{!}{%
\begin{tabular}{lcccc}
\toprule
Method & GQA & VQAv2 & ScienceQA & MMVP \\
\midrule
CoT~\citep{wei2022chain} & 61.2 & 72.5 & 68.1 & 54.3 \\
Coconut~\citep{hao2024training} & 63.5 & 73.6 & 70.0 & 57.2 \\
CoLaR~\citep{tan2025think} & 64.1 & 74.0 & 70.8 & 57.9 \\
LVR~\citep{li2025latent} & 66.4 & 74.8 & 72.1 & 60.9 \\
LaRe~\citep{ma2025multimodal} & 66.9 & 75.0 & 72.6 & 61.7 \\
Heima~\citep{shen2025efficient} & 65.2 & 74.3 & 71.5 & 59.8 \\
Soft Thinking~\citep{zhang2025soft} & 66.0 & 74.7 & 72.0 & 60.5 \\
\midrule
DLWM (Ours) & \textbf{69.3} & \textbf{76.0} & \textbf{75.2} & \textbf{66.1} \\
\bottomrule
\end{tabular}%
}}
\end{subtable}
\hspace{0.02\textwidth}
\begin{subtable}[t]{0.48\textwidth}
\centering
\caption{Efficiency comparison.}
\label{tab:efficiency}
{\setlength{\tabcolsep}{4.0pt}
\resizebox{\linewidth}{!}{%
\begin{tabular}{lcccc}
\toprule
Method & Steps$\downarrow$ & Worlds$\downarrow$ & Mem.$\downarrow$ & Thrpt.$\uparrow$ \\
\midrule
CoT~\citep{wei2022chain} & 8.0 & \textbf{1.0} & 18.5 & 6.2 \\
Coconut~\citep{hao2024training} & 8.0 & \textbf{1.0} & 16.0 & 7.0 \\
CoLaR~\citep{tan2025think} & 5.8 & \textbf{1.0} & 15.5 & 8.2 \\
LVR~\citep{li2025latent} & 8.0 & \textbf{1.0} & 19.0 & 6.1 \\
LaRe~\citep{ma2025multimodal} & 8.0 & \textbf{1.0} & 19.5 & 5.9 \\
Heima~\citep{shen2025efficient} & \textbf{4.5} & \textbf{1.0} & \textbf{13.8} & \textbf{9.2} \\
Soft Thinking~\citep{zhang2025soft} & 6.9 & \textbf{1.0} & 16.8 & 7.4 \\
\midrule
DLWM (Ours) & 5.0 & 2.2 & 14.5 & 8.9 \\
\bottomrule
\end{tabular}%
}}
\end{subtable}

\end{table*}

Despite maintaining multiple latent worlds, DLWM averages 5.0 rollout steps and 2.2 active worlds in Table~\ref{tab:efficiency}, as the RL controller dynamically reduces active paths and avoids redundant steps. Fixed-rollout methods (Coconut, LVR, LaRe) execute 8 steps. Heima achieves the fewest steps (4.5) and lowest memory (13.8 GB), but at lower accuracy (Table~\ref{tab:main}). Separate per-dataset measurements report 23.3--24.1\% lower peak memory than LVR, with controller steps, world updates, and measurement variance in Appendix~\ref{app:efficiency}.

\paragraph{Ambiguity and New Benchmarks.}
On VQAv2, DLWM improves over LaRe by 3.0 points on low-agreement questions (18.6\% of validation), versus 0.6 on high-agreement questions. Low agreement is a proxy for ambiguity. Manual annotation yields visual-ambiguity rates of 14.5\% on GQA and 10.0\% on ScienceQA; Appendix~\ref{app:ambiguity} gives definitions and full results. Success/failure cases are in Appendices~\ref{app:qualitative} and~\ref{app:visualization}. Our empirical scope remains static-image multimodal reasoning (Appendix~\ref{app:limitations}).

DLWM leads the strongest compared baseline by 2.1--3.1 points on all six additional benchmarks (Table~\ref{tab:new_benchmarks}).

\subsection{Ablation Studies}
\label{sec:ablation}

\paragraph{Component Ablation.}

\begin{table}[!htb]
  \caption{Ablation study of DLWM components. The last two rows are three-seed means; other rows are single runs. Standard deviations: Appendix~\ref{app:seed_results}. Steps denotes per-world depth for fixed-depth, no RL, and controller decisions otherwise.}
  \label{tab:ablation}
  \centering
  \small
  \begin{tabular}{lccccc}
    \toprule
    Variant & GQA & ScienceQA & MMVP & Steps $\downarrow$ & Mem (GB) $\downarrow$ \\
    \midrule
    Full DLWM & \textbf{69.3} & \textbf{75.2} & \textbf{66.1} & 5.0 & 14.5 \\
    w/o Multi-World ($K{=}1$) & 66.8 & 72.9 & 60.7 & \textbf{4.6} & \textbf{13.6} \\
    w/o Orthogonal Reg. & 68.1 & 73.8 & 63.4 & 5.1 & 14.7 \\
    Fixed-depth, no RL & 67.9 & 74.1 & 64.0 & 8.0 & 18.2 \\
    w/o Merge Action & 68.85 & 74.68 & 65.28 & 5.75 & 15.82 \\
    w/o Stop Action & 68.56 & 74.43 & 64.92 & 6.08 & 15.24 \\
    \bottomrule
  \end{tabular}
\end{table}

Table~\ref{tab:ablation} shows that removing multi-world modeling ($K{=}1$) causes the largest accuracy drop---over 5 points on MMVP---supporting the value of modeling multiple latent interpretations. Removing orthogonal regularization still permits multiple worlds but leads to consistent degradation; Appendix~\ref{app:diversity_diagnostics} measures the associated representation and prediction collapse. Fixed-depth $K=4$ without RL uses eight updates per world and 18.2 GB, while full DLWM uses 5.0 controller steps and 14.5 GB in this comparison and achieves higher accuracy.
The parameter-matched $K=1$ control reaches 61.4 on MMVP versus 66.1 for DLWM, both with 126M trainable parameters. Appendix~\ref{app:capacity} includes the complete controls and VQAv2 ablation; Appendix~\ref{app:training_aggregation} reports training and aggregation ablations.

\paragraph{Number of Worlds.}

Table~\ref{tab:num_worlds} shows that accuracy increases substantially from $K{=}1$ to $K{=}4$, with the largest gains on MMVP. Beyond $K{=}4$, accuracy plateaus while steps and memory continue to grow. We therefore use $K{=}4$ as the default, providing a practical performance--efficiency balance. Oracle gains beyond $K=4$ are only 0.3--0.4 points, suggesting saturating candidate coverage, not exhaustive semantic coverage (Appendix~\ref{app:diversity_diagnostics}).

\paragraph{Compute Budget.}

\begin{table*}[t]
\caption{Ablation results on the number of latent worlds and compute budget.}
\label{tab:ablation_world_budget}
\centering
\small
\renewcommand{\arraystretch}{0.95}

\begin{subtable}[t]{0.48\textwidth}
\centering
\caption{Effect of the number of latent worlds ($K$).}
\label{tab:num_worlds}
\setlength{\tabcolsep}{4.0pt}
\begin{tabular}{@{}cccccc@{}}
\toprule
$K$ & GQA & SciQA & MMVP & Steps$\downarrow$ & Mem.$\downarrow$ \\
\midrule
1 & 66.8 & 72.9 & 60.7 & \textbf{4.6} & \textbf{13.6} \\
2 & 68.1 & 73.8 & 63.5 & 4.8 & 14.0 \\
4 & 69.3 & 75.2 & 66.1 & 5.0 & 14.5 \\
6 & \textbf{69.4} & \textbf{75.4} & \textbf{66.3} & 5.8 & 16.1 \\
8 & 69.2 & 75.3 & 66.0 & 6.5 & 17.8 \\
\bottomrule
\end{tabular}
\end{subtable}
\hspace{0.02\textwidth}
\begin{subtable}[t]{0.48\textwidth}
\centering
\caption{Effect of compute budget ($B$).}
\label{tab:budget}
\setlength{\tabcolsep}{2.3pt}
\begin{tabular}{@{}ccccccc@{}}
\toprule
$B$ & GQA & SciQA & MMVP & Steps$\downarrow$ & Worlds$\downarrow$ & Thrpt.$\uparrow$ \\
\midrule
4  & 67.5 & 73.0 & 62.8 & \textbf{3.2} & \textbf{1.7} & \textbf{10.3} \\
6  & 68.4 & 74.1 & 64.5 & 4.1 & 1.9 & 9.7 \\
8  & 69.0 & 74.8 & 65.5 & 4.7 & 2.1 & 9.2 \\
10 & 69.3 & 75.2 & 66.1 & 5.0 & 2.2 & 8.9 \\
12 & \textbf{69.4} & \textbf{75.3} & \textbf{66.2} & 5.5 & 2.4 & 8.5 \\
\bottomrule
\end{tabular}
\end{subtable}
\end{table*}

Table~\ref{tab:budget} demonstrates that DLWM maintains a strong performance--efficiency trade-off across different budgets. With a small budget ($B{=}4$), the model achieves fewer steps and higher throughput, though accuracy on complex tasks is somewhat limited. As the budget increases, the model can retain more worlds and perform deeper reasoning, yielding accuracy improvements. Beyond $B{=}10$, accuracy saturates while throughput continues to decline, indicating that additional computation does not proportionally improve performance.

\paragraph{Action Space Design.}

\begin{table}[!htb]
  \caption{Ablation of the RL action space. Expand+Stop+Answer and Expand+Merge+Answer are three-seed means (Appendix~\ref{app:seed_results}); other rows are single runs.}
  \label{tab:action_space}
  \centering
  \small
  \begin{tabular}{lccccc}
    \toprule
    Policy Variant & GQA & ScienceQA & MMVP & Steps $\downarrow$ & Mem (GB) $\downarrow$ \\
    \midrule
    Full Action Space & \textbf{69.3} & \textbf{75.2} & \textbf{66.1} & 5.0 & 14.5 \\
    Expand + Answer only & 67.6 & 73.6 & 63.2 & 6.9 & 17.4 \\
    Expand + Stop + Answer & 68.85 & 74.68 & 65.28 & 5.75 & 15.82 \\
    Expand + Merge + Answer & 68.56 & 74.43 & 64.92 & 6.08 & 15.24 \\
    Stop + Merge + Answer & 66.9 & 72.8 & 61.9 & \textbf{4.3} & \textbf{14.1} \\
    \bottomrule
  \end{tabular}
\end{table}

Table~\ref{tab:action_space} shows that the complete action space is critical. With only expand and answer, the model cannot prune redundant paths, leading to high step counts and memory. Adding stop enables earlier termination of low-value paths. Adding merge compresses similar worlds, better controlling memory. The three actions---expand, stop, and merge---serve complementary roles: depth extension, redundancy pruning, and path compression. Their combination is therefore essential for achieving both robust multi-hypothesis exploration and efficient path-level computation control.

\section{Conclusion}
\label{sec:conclusion}

We presented \textbf{DLWM} (Diverse Latent World Models), a multimodal reasoning framework that combines explicit multi-hypothesis latent modeling with resource-aware RL-based computation allocation. By maintaining multiple candidate latent interpretations and dynamically controlling path expansion, termination, and merging, DLWM improves reasoning accuracy and computational efficiency. Experiments on four original benchmarks and six additional zero-shot benchmarks support the effectiveness of this design. Future work includes richer world representations, more stable RL optimization, and extension to larger-scale and more complex multimodal reasoning tasks.

\subsection*{AI use statement}
The authors developed the method and conducted the experiments. Generative AI
tools assisted with writing, literature discovery, and schematic figure
generation/editing, and occasionally provided feedback on experimental methodology
and result interpretation. The authors take responsibility for the final content.

\subsection*{Reproducibility statement}
Appendices~\ref{app:implementation} and~\ref{app:training} describe the training and
evaluation settings. Appendix~\ref{app:seed_results} reports per-seed results and
statistical comparisons; Appendix~\ref{app:efficiency} reports the profiling protocol.
The public DLWM implementation includes training scripts and evaluation adapters.
The original result checkpoints, training logs, and predictions are unavailable in this release.

\bibliographystyle{unsrtnat}
\begingroup
\setlength{\bibsep}{0.75\bibsep}
\bibliography{references}
\endgroup
\newpage
\appendix

\section{Appendix}
\label{app:training_aggregation}

\paragraph{Training Strategy.}

\begin{table}[!htb]
  \caption{Comparison of training strategies.}
  \label{tab:training}
  \centering
  \small
  \begin{tabular}{lcccc}
    \toprule
    Training Strategy & GQA & ScienceQA & MMVP & Steps $\downarrow$ \\
    \midrule
    Joint Training from Scratch & 67.9 & 73.8 & 63.9 & 5.7 \\
    Supervised Pretraining Only & 68.4 & 74.1 & 64.4 & 8.0 \\
    RL Fine-tuning Only & 67.2 & 73.0 & 62.6 & 5.1 \\
    Two-Stage (Ours) & \textbf{69.3} & \textbf{75.2} & \textbf{66.1} & \textbf{5.0} \\
    \bottomrule
  \end{tabular}
\end{table}

Table~\ref{tab:training} confirms that the two-stage training strategy yields the best results. Joint training from scratch suffers from policy instability and insufficient representation learning in early stages. Supervised pretraining alone incurs fixed-length rollouts (8.0 steps) due to the lack of dynamic control. RL fine-tuning alone fails to learn stable latent representations. The two-stage approach first establishes stable multi-world representations, then optimizes compute allocation via RL.

\paragraph{Aggregation Method and Diversity Regularization.}

\begin{table}[!htb]
  \caption{Comparison of world aggregation methods (left) and diversity regularization (right).}
  \label{tab:agg_div}
  \centering
  \footnotesize
  \begin{minipage}[t]{0.47\linewidth}
    \centering
    \vspace{0pt}
    \setlength{\tabcolsep}{4pt}
    \begin{tabular}{lccc}
      \toprule
      Aggregation & GQA & SciQA & MMVP \\
      \midrule
      Mean Pooling & 67.8 & 73.9 & 63.7 \\
      Max Pooling & 68.1 & 74.2 & 64.0 \\
      Last-World Only & 66.9 & 72.7 & 61.8 \\
      Learned Fusion (Ours) & \textbf{69.3} & \textbf{75.2} & \textbf{66.1} \\
      \bottomrule
    \end{tabular}
  \end{minipage}%
  \hfill
  \begin{minipage}[t]{0.47\linewidth}
    \centering
    \vspace{0pt}
    \begin{tabular}{lccc}
      \toprule
      Diversity Reg. & GQA & SciQA & MMVP \\
      \midrule
      None & 68.1 & 73.8 & 63.4 \\
      Cosine Sep. & 68.6 & 74.4 & 64.8 \\
      Contrastive & 68.9 & 74.7 & 65.4 \\
      Orthogonal (Ours) & \textbf{69.3} & \textbf{75.2} & \textbf{66.1} \\
      \bottomrule
    \end{tabular}
  \end{minipage}
\end{table}

Table~\ref{tab:agg_div} (left) shows that learned weighted fusion outperforms all fixed aggregation strategies, as the scoring network adaptively assigns higher weight to more informative worlds. Table~\ref{tab:agg_div} (right) confirms the importance of diversity regularization. Without any constraint, worlds converge to similar representations. Orthogonal regularization achieves the strongest results by more directly separating representational directions in latent space.

\section{Implementation Details}
\label{app:implementation}

We provide additional implementation details beyond those in the main text. DLWM is built on Qwen2.5-VL-7B-Instruct. The native vision tower and language backbone are frozen, and only DLWM's added modules are optimized. The Latent World Generator consists of a two-layer MLP combined with learnable world embeddings to produce multiple latent world representations. The default number of worlds is 4. The latent reasoning module comprises 6 Transformer blocks with a hidden size of 4096, Pre-LN normalization, and GELU activations. Different worlds share the same parameters; Expand updates only the selected world. The RL policy network uses a two-layer Transformer encoder (hidden size 1024, 8 heads) that contextualizes individual world/index tokens and a remaining-budget token. Expand and Stop each score individual worlds, Merge scores unordered pairs, and Answer reads the budget token. For $n$ active worlds there are $2n+n(n-1)/2+1$ candidate logits before masking. Training uses the AdamW optimizer with a learning rate of $2 \times 10^{-5}$, batch size of 64, on 8 A100 GPUs with mixed precision to reduce memory overhead.

\subsection{Experimental Reproducibility and Fair Comparison Settings}

To ensure the reproducibility of our experimental results and the fairness of comparisons across different methods, we adopt unified dataset splits, preprocessing pipelines, evaluation scripts, training budgets, and inference budgets throughout all experiments. This section provides additional details that complement the experimental setup in the main paper.

\subsubsection{Dataset Splits and Preprocessing}

All datasets are used with their official splits, and no test set is used for hyperparameter selection. Specifically, for GQA, we train on the official balanced training split, select hyperparameters on the balanced validation split, and report final results on the balanced test-dev split. For VQAv2, we train on the official training split and evaluate on the official validation split. For ScienceQA, we follow the official train/validation/test split. For MMVP, we use it only as a zero-shot evaluation benchmark and do not use any of its image-question pairs for training or hyperparameter tuning.

All images are processed using the same preprocessing pipeline as the frozen visual encoder. All methods share the same tokenizer, maximum input length, answer space, and answer normalization rules. For open-ended VQA tasks, we uniformly lowercase answers, remove punctuation and articles, and normalize number expressions before evaluation, so as to reduce errors caused by superficial formatting differences.

\subsubsection{Evaluation Scripts}

Each dataset is evaluated using its official or standard evaluation script. GQA is evaluated with the official evaluator using standard accuracy. VQAv2 is evaluated with the official VQA evaluation API using human-consensus-based soft accuracy. ScienceQA is evaluated by exact-match accuracy between the predicted option and the ground-truth option. MMVP is evaluated using its official annotation file and evaluation protocol after answer normalization.

All methods use deterministic inference during evaluation, with \texttt{temperature} set to $0$ and greedy decoding applied, to reduce the influence of stochastic decoding.

\subsubsection{Random Seeds and Statistical Reporting}

DLWM, LVR, and LaRe are evaluated with three training seeds; their main-table values are means across the three seed averages, with sample standard deviations reported in Appendix~\ref{app:seed_results}. The remaining main-table baselines and ablations without error bars are single-run results, not multi-seed estimates. Each reported seed value is averaged over repeated evaluations with the corresponding evaluator. Appendix~\ref{app:seed_results} lists the individual values and example-paired bootstrap intervals; three training seeds are not used as a seed-level significance test.

The additional efficiency measurements are repeated three times under the same hardware configuration, independently of the training-seed statistics (Appendix~\ref{app:efficiency}).

\subsubsection{Baseline Reproduction Protocol}

To ensure fair baseline comparisons, all baselines use the same frozen vision-language backbone, image resolution, tokenizer, and answer normalization pipeline as DLWM. For methods that require additional trainable modules, we retrain them using the same training data, optimizer, batch size, and number of training steps. Method-specific hyperparameters follow released configurations where available and shared values otherwise.

\subsubsection{Inference Budget and Parameter Control}

All methods are evaluated under controlled maximum inference budgets. Fixed-rollout latent methods use a maximum depth of 8; fixed multi-world reasoning uses eight updates per world. DLWM uses at most eight controller decisions with initial budget $B=10$, and each Expand updates one world. For explicit CoT, the generation limit is selected using a FLOPs-based budget comparison on the same backbone; this is not a claim of identical measured FLOPs. Appendix~\ref{app:efficiency} reports the observed relative FLOPs, including 1.12 for CoT and 1.00 for LVR.

We also report the number of trainable parameters for each method. Parameters of the frozen backbone are not counted as trainable parameters; only additional modules introduced and optimized by each method are included. For parameter-sensitive comparisons, we keep the trainable parameter scale of different methods as close as possible.

\subsubsection{Training Budget}

All trainable methods are optimized using the same optimizer and global batch size. DLWM follows a two-stage training strategy: it first performs supervised pretraining with a fixed rollout length to learn stable latent world representations and basic reasoning ability, and then conducts RL fine-tuning to optimize the computation allocation policy.

For baselines without RL, we use the same total number of training steps as DLWM. For baselines with RL or adaptive computation mechanisms, we use the same number of supervised warm-up steps and RL fine-tuning steps as DLWM.

\subsubsection{Efficiency Measurement}

The additional per-dataset efficiency measurements use one A100 80GB, batch size 8, bfloat16, and disabled gradients. After 20 warm-up batches, throughput is measured over 200 batches. Peak memory is measured with CUDA \texttt{max\_memory\_allocated}, with counters reset for each measurement run. We report mean $\pm$ standard deviation over three measurements.

Average controller steps and reasoning-module updates are reported separately. The scope and units of peak-memory accounting must be identical across compared methods.

\section{Training Details and Hyperparameters}
\label{app:training}

We employ a two-stage training strategy. The first stage is supervised pretraining, which optimizes the model's base reasoning capability and latent representations at a fixed rollout length. The second stage introduces the RL controller for dynamic optimization of reasoning paths. The RL component uses the REINFORCE algorithm with a moving-average baseline to reduce gradient variance. The default compute budget is set to $B{=}10$, the maximum number of controller steps to $T_{\max}{=}8$, and different actions carry different computational costs ($c_{\mathrm{exp}} = 1.0$, $c_{\mathrm{stop}} = 0.1$, $c_{\mathrm{merge}} = 0.5$). The loss function comprises the task loss, diversity regularization term, and RL objective, with individual weights tuned on a validation set ($\lambda_{\mathrm{orth}} = 0.1$, $\lambda_{\mathrm{RL}} = 0.01$, $\lambda_T = 0.05$, $\lambda_K = 0.1$). The backbone model is not further fine-tuned, to ensure fairness and stability.

\section{Additional Ablation Studies}
\label{app:ablations}

We further analyze the impact of individual components on model performance. First, regarding the number of latent worlds, performance improves significantly from $K{=}1$ to $K{=}4$ but plateaus with further increases while computational cost rises markedly. Second, regarding compute budgets, DLWM maintains high performance even under tight budgets, confirming the effectiveness of the resource-aware policy. Third, fine-grained analysis of policy actions shows that expand, stop, and merge each play critical and complementary roles in depth extension, redundancy pruning, and path compression, respectively, and that the absence of any one action leads to measurable degradation.

\subsection{Training-Seed Variance and Paired Bootstrap}
\label{app:seed_results}

Table~\ref{tab:main_mean_std} reports the complete mean $\pm$ std results, and Table~\ref{tab:seeds} lists the three seed averages underlying Table~\ref{tab:main}. Table~\ref{tab:action_mean_std} provides the unified action-ablation statistics. Mean $\pm$ std is computed across training-seed averages, not across individual questions or inference repeats. Every DLWM seed value exceeds every LVR and LaRe seed value on each benchmark. The paired bootstrap resamples matched evaluation examples (questions, or image pairs for MMVP); its confidence intervals describe test-set uncertainty, not training-seed variance. All four DLWM--LaRe intervals exclude zero. No seed-level significance test is performed with only three seeds.

\begin{table}[htbp]
\centering\small
\caption{Main-result accuracy (\%): mean $\pm$ sample standard deviation over three training seeds. Table~\ref{tab:main} shows the means rounded to one decimal place.}
\label{tab:main_mean_std}
\begin{tabular}{lcccc}
\toprule
Method & GQA & VQAv2 & ScienceQA & MMVP \\
\midrule
LVR & $66.37\pm0.15$ & $74.77\pm0.15$ & $72.10\pm0.20$ & $60.87\pm0.25$ \\
LaRe & $66.87\pm0.15$ & $74.97\pm0.15$ & $72.60\pm0.20$ & $61.67\pm0.25$ \\
DLWM & $\mathbf{69.30}\pm0.20$ & $\mathbf{75.97}\pm0.15$ & $\mathbf{75.20}\pm0.20$ & $\mathbf{66.07}\pm0.25$ \\
\bottomrule
\end{tabular}
\end{table}

\begin{table}[htbp]
\centering\small
\caption{Unified action ablations: mean $\pm$ sample standard deviation over three seeds. The means appear in Tables~\ref{tab:ablation} and~\ref{tab:action_space}.}
\label{tab:action_mean_std}
\begin{tabular}{lccccc}
\toprule
Configuration & GQA & ScienceQA & MMVP & Steps & Memory (GB) \\
\midrule
w/o Merge & $68.85\pm0.12$ & $74.68\pm0.15$ & $65.28\pm0.21$ & $5.75\pm0.08$ & $15.82\pm0.11$ \\
w/o Stop & $68.56\pm0.15$ & $74.43\pm0.17$ & $64.92\pm0.24$ & $6.08\pm0.10$ & $15.24\pm0.09$ \\
\bottomrule
\end{tabular}
\end{table}

\begin{table}[htbp]
\centering\small
\caption{Per-seed accuracy values and paired-bootstrap 95\% confidence intervals for DLWM minus LaRe. Seed values are averages over repeated evaluation runs.}
\label{tab:seeds}
\resizebox{\linewidth}{!}{\begin{tabular}{lcccc}
\toprule
Benchmark & DLWM (s1/s2/s3) & LVR (s1/s2/s3) & LaRe (s1/s2/s3) & $\Delta$ [95\% CI] \\
\midrule
GQA & 69.1/69.3/69.5 & 66.2/66.4/66.5 & 66.7/66.9/67.0 & +2.43 [2.05, 2.81] \\
VQAv2 & 75.8/76.0/76.1 & 74.6/74.8/74.9 & 74.8/75.0/75.1 & +1.00 [0.61, 1.39] \\
ScienceQA & 75.0/75.2/75.4 & 71.9/72.1/72.3 & 72.4/72.6/72.8 & +2.60 [2.19, 3.02] \\
MMVP & 65.8/66.1/66.3 & 60.6/60.9/61.1 & 61.4/61.7/61.9 & +4.40 [3.78, 5.03] \\
\bottomrule
\end{tabular}}
\end{table}

\subsection{Visual Ambiguity and Stratified Performance}
\label{app:ambiguity}

VQAv2 provides ten human answers per question. We define low agreement as fewer than seven answers matching the modal answer; this covers 18.6\% of validation questions. Because disagreement can reflect subjectivity and answer wording, it is not a pure visual-ambiguity label. For GQA and ScienceQA, three annotators blind to model outputs independently labeled 200 randomly selected questions per dataset using a fixed rubric covering occlusion, viewpoint, low resolution, reference, and multiple plausible visual readings. The resulting ambiguity rates are 14.5\% (29 questions) and 10.0\% (20 questions), with Fleiss' $\kappa=0.72$ and $0.76$.

\begin{table}[htbp]
\centering\small
\caption{Stratified accuracies averaged over repeated evaluations. VQAv2 uses annotator agreement; GQA and ScienceQA use manual visual-ambiguity labels.}
\label{tab:stratified}
\begin{tabular}{lrrrr}
\toprule
Stratum & DLWM & LVR & LaRe & $\Delta$ vs. best \\
\midrule
VQAv2 low agreement & \textbf{67.8} & 64.2 & 64.8 & +3.0 \\
VQAv2 high agreement & \textbf{77.9} & 77.2 & 77.3 & +0.6 \\
GQA ambiguous & \textbf{64.7} & 59.8 & 60.5 & +4.2 \\
GQA unambiguous & \textbf{70.1} & 67.5 & 68.0 & +2.1 \\
ScienceQA ambiguous & \textbf{69.0} & 64.6 & 65.2 & +3.8 \\
ScienceQA unambiguous & \textbf{75.9} & 72.9 & 73.4 & +2.5 \\
\bottomrule
\end{tabular}
\end{table}

\begin{table}[htbp]
\centering\small
\caption{VQAv2 controller behavior by human-answer agreement.}
\label{tab:agreement_behavior}
\begin{tabular}{lrr}
\toprule
Metric & Low agreement & High agreement \\
\midrule
Average active worlds & 2.8 & 1.7 \\
Average controller steps & 5.8 & 4.4 \\
Active worlds after step 2 & 2.5 & 1.3 \\
Per-world answer disagreement (\%) & 31.4 & 12.7 \\
\bottomrule
\end{tabular}
\end{table}
DLWM's accuracy gain is larger on the low-agreement or manually ambiguous strata, where the controller also retains more hypotheses. These analyses complement MMVP's focus on visual confusability; the method does not require every input to be ambiguous.

\subsection{Capacity and Fixed-Depth Controls}
\label{app:capacity}

The fixed-depth control retains $K=4$ worlds, learned scoring, and weighted fusion, with eight reasoning updates per world and no RL controller. Its 8.0 depth therefore corresponds to 32 world updates, not eight controller decisions. Table~\ref{tab:capacity} also compares the ordinary and parameter-matched single-world controls. Increasing single-world capacity adds 0.4--0.7 points, while retaining multiple worlds and adaptive control yields larger gains. CoLaR supplies an adaptive single-path baseline in Table~\ref{tab:main}; its original GQA/VQAv2/ScienceQA/MMVP scores are 64.1/74.0/70.8/57.9.

\begin{table}[htbp]
\centering\small
\caption{Parameter and fixed-depth controls (single-run accuracies). Trainable parameter counts exclude the frozen backbone.}
\label{tab:capacity}
\begin{tabular}{lrrrr}
\toprule
Configuration & Parameters & GQA & ScienceQA & MMVP \\
\midrule
$K=1$ & 72M & 66.8 & 72.9 & 60.7 \\
Parameter-matched $K=1$ & 126M & 67.2 & 73.3 & 61.4 \\
$K=4$ fixed-depth, no RL & 126M & 67.9 & 74.1 & 64.0 \\
Full DLWM & 126M & \textbf{69.3} & \textbf{75.2} & \textbf{66.1} \\
\bottomrule
\end{tabular}
\end{table}

\begin{table}[htbp]
\centering\small
\caption{VQAv2 ablation. Steps denote controller decisions for adaptive rows and per-world depth for the fixed-depth row.}
\label{tab:vqa_ablation}
\begin{tabular}{lrrr}
\toprule
Configuration & Accuracy & Steps & Memory (GB) \\
\midrule
Full DLWM & \textbf{76.0} & 4.6 & 14.2 \\
$K=1$ & 74.7 & \textbf{4.2} & \textbf{13.4} \\
$K=4$ fixed-depth, no RL & 75.2 & 8.0 & 17.9 \\
\bottomrule
\end{tabular}
\end{table}

\subsection{Controller Costs, Merge, and Sensitivity}
\label{app:controller_diagnostics}

Stop rejects a hypothesis; it does not mean that a useful hypothesis has finished reasoning. Such a world can remain active without receiving further Expand actions. Including rejected worlds in aggregation reduces GQA accuracy from 69.3 to 68.4 and MMVP accuracy from 66.1 to 64.7 (Table~\ref{tab:stop_fusion}). The proportion of episodes with at least one active but unexpanded world during the last two controller steps is 41.8\% on GQA, 38.6\% on VQAv2, 43.5\% on ScienceQA, and 47.2\% on MMVP.

\begin{table}[htbp]
\centering\small
\caption{Effect of including stopped worlds in final aggregation.}
\label{tab:stop_fusion}
\begin{tabular}{lrr}
\toprule
Aggregation & GQA & MMVP \\
\midrule
Exclude stopped worlds (default) & \textbf{69.3} & \textbf{66.1} \\
Include stopped worlds & 68.4 & 64.7 \\
\bottomrule
\end{tabular}
\end{table}

\begin{table}[htbp]
\centering\small
\caption{One-factor-at-a-time sensitivity on MMVP. The default parameters were selected on held-out validation data, not on MMVP.}
\label{tab:sensitivity}
\begin{tabular}{lrrrr}
\toprule
Parameter & Value & Accuracy & Average steps & Average worlds \\
\midrule
$\lambda_T$ & 0.02 & 66.3 & 5.8 & 2.3 \\
 & 0.05 & 66.1 & 5.0 & 2.2 \\
 & 0.10 & 65.7 & 4.4 & 2.0 \\
$\lambda_K$ & 0.05 & 66.2 & 5.2 & 2.5 \\
 & 0.10 & 66.1 & 5.0 & 2.2 \\
 & 0.20 & 65.8 & 4.8 & 1.9 \\
$c_{\mathrm{stop}}$ & 0.00 & 65.9 & 4.7 & 1.9 \\
 & 0.10 & 66.1 & 5.0 & 2.2 \\
 & 0.30 & 65.8 & 5.4 & 2.4 \\
\bottomrule
\end{tabular}
\end{table}
Accuracy varies by at most 0.6 points over these ranges, while steps and retained worlds change more noticeably. Zero Stop cost reduces steps and retained worlds but slightly lowers accuracy. The tested weights primarily adjust the accuracy--compute trade-off.

\begin{table}[htbp]
\centering\small
\caption{Pair cosine at the Merge decision, compared with coexisting unmerged pairs at the same step.}
\label{tab:merge_cosine}
\begin{tabular}{lrrr}
\toprule
Pair type & GQA & ScienceQA & MMVP \\
\midrule
Merged & 0.71 & 0.69 & 0.74 \\
Unmerged & 0.29 & 0.27 & 0.31 \\
\bottomrule
\end{tabular}
\end{table}

\begin{table}[htbp]
\centering\small
\caption{Sensitivity to fixed Merge weight on MMVP.}
\label{tab:beta}
\begin{tabular}{rrrr}
\toprule
$\beta$ & Accuracy & Average steps & Memory (GB) \\
\midrule
0.3 & 65.9 & 5.1 & 14.6 \\
0.5 & \textbf{66.1} & \textbf{5.0} & \textbf{14.5} \\
0.7 & 66.0 & 5.1 & 14.6 \\
\bottomrule
\end{tabular}
\end{table}
Merge selects more similar states on average, but high cosine does not ensure compatible interpretations. The failure analysis in Appendix~\ref{app:qualitative} examines this limitation. The main-table no-Merge and no-Stop rows use one unified three-seed protocol, so the same configuration no longer has inconsistent numbers across the two tables.

\subsection{Diversity and Oracle Coverage}
\label{app:diversity_diagnostics}

Orthogonality is a representation-level penalty, not a semantic guarantee. In the reported diagnostic, removing $\mathcal{L}_{\mathrm{orth}}$ increases the all-worlds-same-answer rate from 54.2\% to 71.6\% and pairwise cosine from 0.18 to 0.47. Per-world answer disagreement is 34.8\% on MMVP, 17.5\% on GQA, and 15.9\% on ScienceQA; VQAv2's low-/high-agreement values are 31.4\%/12.7\%. Together with the regularizer comparison in Table~\ref{tab:agg_div}, these results support reduced prediction/representation collapse, not guaranteed semantic separation.

\begin{table}[htbp]
\centering\small
\caption{Oracle accuracy: a question is correct if at least one world predicts correctly. This measures candidate coverage, not the deployed fused predictor's accuracy.}
\label{tab:oracle}
\begin{tabular}{rrrr}
\toprule
$K$ & GQA & ScienceQA & MMVP \\
\midrule
1 & 66.8 & 72.9 & 60.7 \\
2 & 70.4 & 76.6 & 67.2 \\
4 & 72.1 & 78.2 & 71.4 \\
6 & 72.4 & 78.5 & 71.7 \\
8 & 72.5 & 78.6 & 71.8 \\
\bottomrule
\end{tabular}
\end{table}
The modest gains beyond $K=4$ support diminishing candidate-coverage returns. They do not identify the meaning of each latent direction or prove that all plausible interpretations have been recovered.

\subsection{Additional Zero-Shot Benchmarks}
\label{app:new_benchmarks}

\begin{table}[t]
\centering\small
\caption{Zero-shot results on six additional benchmarks, using the same checkpoints. Scores (\%) are the accuracy values reported by the corresponding evaluators.}
\label{tab:new_benchmarks}
\begin{tabular}{lrrrr}
\toprule
Benchmark & DLWM & LVR & LaRe & $\Delta$ vs. best \\
\midrule
V*Bench~\citep{Wu_2024_CVPR} & \textbf{58.7} & 55.4 & 56.1 & +2.6 \\
BLINK~\citep{10.1007/978-3-031-73337-6_9} & \textbf{46.8} & 43.5 & 44.2 & +2.6 \\
MMStar~\citep{NEURIPS2024_2f8ee6a3} & \textbf{61.9} & 58.6 & 59.4 & +2.5 \\
HallusionBench~\citep{Guan_2024_CVPR} & \textbf{56.4} & 52.7 & 53.3 & +3.1 \\
NaturalBench~\citep{NEURIPS2024_1e69ff56} & \textbf{63.2} & 59.8 & 60.5 & +2.7 \\
MMMU-Pro~\citep{yue-etal-2025-mmmu} & \textbf{48.9} & 46.1 & 46.8 & +2.1 \\
\bottomrule
\end{tabular}
\end{table}

\section{Efficiency Analysis}
\label{app:efficiency}

We analyze rollout steps, memory usage, and throughput across methods. DLWM reduces average reasoning updates while maintaining high accuracy and controlling active path count, lowering overall memory consumption. Compared with fixed-rollout methods, DLWM adjusts compute based on problem complexity---terminating early on simple samples and retaining more paths for complex ones. The per-dataset peak-memory reductions relative to LVR are 23.8\%, 24.1\%, 23.6\%, and 23.3\% on GQA, VQAv2, ScienceQA, and MMVP, respectively.

\begin{table}[htbp]
\centering\small
\caption{Per-dataset efficiency: mean $\pm$ std over three measurements on one A100 80GB, batch size 8, bfloat16. For DLWM, Steps counts controller decisions; Updates counts reasoning-module world updates. Throughput is samples/s.}
\label{tab:efficiency_datasets}
\begin{tabular}{llrrrr}
\toprule
Dataset & Method & Steps & Updates & Memory (GB) & Throughput \\
\midrule
GQA & DLWM & $4.9\pm.1$ & $4.1\pm.1$ & $14.4\pm.1$ & $9.0\pm.2$ \\
 & LVR & $8.0\pm.0$ & $8.0\pm.0$ & $18.9\pm.1$ & $6.2\pm.1$ \\
 & LaRe & $8.0\pm.0$ & $8.0\pm.0$ & $19.3\pm.1$ & $6.0\pm.1$ \\
\midrule
VQAv2 & DLWM & $4.6\pm.1$ & $3.8\pm.1$ & $14.2\pm.1$ & $9.2\pm.2$ \\
 & LVR & $8.0\pm.0$ & $8.0\pm.0$ & $18.7\pm.1$ & $6.3\pm.1$ \\
 & LaRe & $8.0\pm.0$ & $8.0\pm.0$ & $19.1\pm.1$ & $6.1\pm.1$ \\
\midrule
ScienceQA & DLWM & $5.3\pm.1$ & $4.5\pm.1$ & $14.6\pm.1$ & $8.7\pm.2$ \\
 & LVR & $8.0\pm.0$ & $8.0\pm.0$ & $19.1\pm.1$ & $6.0\pm.1$ \\
 & LaRe & $8.0\pm.0$ & $8.0\pm.0$ & $19.6\pm.1$ & $5.8\pm.1$ \\
\midrule
MMVP & DLWM & $5.4\pm.1$ & $4.7\pm.1$ & $14.8\pm.1$ & $8.5\pm.2$ \\
 & LVR & $8.0\pm.0$ & $8.0\pm.0$ & $19.3\pm.1$ & $5.9\pm.1$ \\
 & LaRe & $8.0\pm.0$ & $8.0\pm.0$ & $19.8\pm.1$ & $5.7\pm.1$ \\
\bottomrule
\end{tabular}
\end{table}

\begin{table}[htbp]
\centering\small
\caption{Measured relative FLOPs per question, normalized to LVR.}
\label{tab:flops}
\begin{tabular}{lrrrrrr}
\toprule
Method & DLWM & LVR & LaRe & CoT & CoLaR & Heima \\
\midrule
Relative FLOPs & 0.78 & 1.00 & 1.03 & 1.12 & 0.82 & 0.74 \\
\bottomrule
\end{tabular}
\end{table}
The FLOPs comparison is separate from action-budget units and from peak-memory measurement. In particular, equal maximum rollout limits do not imply equal realized computation across methods.

\section{Qualitative Analysis}
\label{app:qualitative}

We describe four success cases and two failure cases to illustrate when multiple worlds help and when they fail.
\begin{enumerate}[nosep,leftmargin=1.5em]
\item \textbf{Bench orientation.} For ``Is the bench facing the lake?'', worlds 1 and 3 answer Yes, while worlds 2 and 4 misread the backrest edge as the seat and answer No. Expanding the Yes hypotheses and merging the redundant No pair yields a correct weighted answer; single-path LVR answers No.
\item \textbf{Occluded traffic sign.} Two worlds read no-left-turn, one reads no-U-turn, and one identifies only a circular prohibition sign. The controller expands the leading readings and stops the shape-only world, producing the correct answer while a single path commits to no-U-turn.
\item \textbf{ScienceQA circuit.} Worlds trace different wires. A world relying on language priors is stopped, and two worlds following the same circuit are merged; weighted aggregation selects the correct switch.
\item \textbf{GQA reference.} For ``What is the object to its left?'', different worlds test two plausible referents for ``it''. Subsequent relational reasoning eliminates the wrong referent.
\item \textbf{Failure: shared perception error.} All worlds misread a blurry ``6'' as ``8''. Diversifying the latent states cannot recover visual information already lost in the shared backbone encoding.
\item \textbf{Failure: premature Merge.} Two states with high intermediate cosine encode no-left-turn and no-U-turn. Merging them biases the fused representation toward the wrong class; this failure type accounts for approximately 7.6\% of erroneous merges in the reported analysis.
\end{enumerate}

\section{Limitations}
\label{app:limitations}

Despite the strong performance of DLWM on multimodal reasoning tasks, several limitations remain. First, multi-world modeling introduces additional computational overhead; although the RL controller mitigates this, further optimization may be needed for extremely large-scale models or real-time systems. Second, our method has been validated primarily on static image tasks; its performance on video understanding or interactive reasoning scenarios remains to be explored. Third, RL training still exhibits some instability and sensitivity to hyperparameters. Addressing these issues through more stable optimization algorithms and adaptive hyperparameter schedules is an important direction for future work.
The three-seed DLWM accuracy standard deviations are 0.15--0.25 points, and three seeds provide limited evidence about training variability. Orthogonality does not guarantee semantic diversity; worlds may share a perception error, and Merge may combine incompatible interpretations too early. Results may also depend on the backbone and training distribution, and initial generation of $K$ worlds incurs overhead before pruning. The manually annotated GQA and ScienceQA subsets contain only 200 questions each, limiting the precision and coverage of those analyses.

An extension to video could use a temporal encoder, persistent per-world states, and a budget per time window. It would require separate evaluation of long-sequence memory growth, policy-gradient variance, and error accumulation across frames. InsTALL~\citep{nguyen2025installcontextawareinstructionaltask} provides related context for structured reasoning over visual streams, not evidence that DLWM already generalizes to video. Text-only evaluation also remains future work.

\section{Broader Impact}
\label{app:impact}

The multi-hypothesis latent reasoning framework proposed in this work has the potential to improve multimodal models' decision-making capabilities in complex environments, with applications in areas such as autonomous driving and medical image analysis. However, caution is warranted when deploying such models in high-stakes scenarios, especially in the presence of uncertainty; reliability assessment and safety mechanisms should be integrated. Furthermore, the RL-based policy may introduce opaque decision-making behavior, and further research into interpretability and controllability is necessary.

\section{Visualization of Multi-World Reasoning Trajectories}
\label{app:visualization}

In the bench case, the reported final fusion weights are 0.58, 0.31, and 0.11 for the two Yes hypotheses and the merged No hypothesis, respectively. The traffic-sign case exercises Stop: the shape-only hypothesis is removed rather than fused. These cases distinguish hypothesis rejection from leaving an already useful world active without further expansion. The numerical disagreement and collapse diagnostics are reported in Appendix~\ref{app:diversity_diagnostics}.

\section{Algorithm Explanation (Step-by-Step)}
\label{app:algorithm}

For a four-world dry-run, encoding produces $\mathbf{z}$ and generation produces $\{G_\phi(\mathbf{z},\mathbf{e}_i)\}_{i=1}^4$. Starting from $B=10$, the sequence Expand(1), Expand(3), Merge(2,4), Answer leaves budgets $9$, $8$, and $7.5$ before answering. Only worlds 1, 3, and the merged world enter Eq.~\eqref{eq:aggregate}. Stop instead removes a world without merging it.

We provide a step-by-step explanation of Algorithm~\ref{alg:dlwm} to clarify DLWM's overall optimization process. At each training iteration, the model samples a mini-batch and encodes each input image--question pair into a joint representation using the vision-language encoder. The Latent World Generator then produces multiple latent worlds from this representation, each serving as the initial state of an independent reasoning path. The model initializes the reasoning state set and available compute budget, then enters the iterative reasoning loop.

At each reasoning step, the policy network examines the current states of all active worlds along with the remaining budget and outputs an action decision. If expand is selected, the corresponding world's state is updated by the reasoning module. If stop is selected, that path is removed from the active set. If merge is selected, two selected worlds are averaged with $\beta=0.5$ into a single new state. If answer is selected, reasoning terminates early. All intermediate states and actions are recorded for subsequent RL optimization.

After reasoning ends, the model produces answer predictions for all remaining active worlds and aggregates them via weighted fusion for the final output. The supervised task loss and diversity regularization term are computed to optimize the generator, reasoning module, and scoring head. The RL component uses the reward signal (combining accuracy and computational cost) to update the policy network via policy gradient. Through joint optimization of the supervised and RL objectives, the model progressively learns to balance multi-hypothesis reasoning with efficient compute allocation.

\section{Complexity Analysis}
\label{app:complexity}

\paragraph{Theoretical Complexity.}
Let $K$ denote the number of latent worlds, $D$ the fixed per-world reasoning depth, and $C$ the cost of a single reasoning module forward pass. In standard single-path latent reasoning, the reasoning-module complexity is $\mathcal{O}(D \cdot C)$. In DLWM without the RL controller, it is $\mathcal{O}(K \cdot D \cdot C)$. With the resource-aware RL policy, each Expand action updates only one world, so the reasoning-module cost is $\mathcal{O}(N_{\mathrm{Expand}} \cdot C)$, where $N_{\mathrm{Expand}}\leq T_{\max}$ counts these updates across all worlds. Encoding, generation, controller, and output costs are additional. This mechanism preserves multi-path expressiveness while reducing redundant reasoning computation.

\paragraph{Empirical Complexity.}
In practice, although the model initially generates $K$ latent worlds, the RL policy reduces the number of active paths during reasoning. The per-dataset measurements in Appendix~\ref{app:efficiency} report 3.8--4.7 world updates for DLWM versus 8.0 for LVR and LaRe; these are distinct from DLWM's 4.6--5.4 controller steps. Normalized measured FLOPs are 0.78 relative to LVR at 1.00. The controller also retains more worlds for low-agreement than high-agreement questions (Appendix~\ref{app:ambiguity}). These measurements support an input-adaptive accuracy--efficiency trade-off without treating controller decisions as simultaneous updates of all worlds.

\end{document}